\documentclass[conference]{IEEEtran}
\usepackage{times}
\IEEEoverridecommandlockouts  

\usepackage[numbers,compress]{natbib}
\usepackage{multicol}
\usepackage[bookmarks=true]{hyperref}

\usepackage{graphicx}
\usepackage{amsfonts}       
\usepackage{multirow}
\usepackage{subcaption}
\usepackage{pgfplots}
\usepackage{pgfplotstable}
\pgfplotsset{compat=1.17}
\usepackage{multicol}
\usepackage{amsmath}
\usepackage[export]{adjustbox}
\usepackage{rotating}
\usepackage{standalone}
\usepackage{booktabs, makecell, tabularx}

\usepackage[export]{adjustbox}
\usepackage{rotating}
\usepackage{standalone}
\usepackage{booktabs, makecell, tabularx}
\usepackage{cleveref}
\usepackage{overpic}

\newcolumntype{s}{>{\hsize=.05\hsize}X}

\usepackage{mathtools}
\DeclarePairedDelimiterX{\infdivx}[2]{(}{)}{%
  #1\;\delimsize\|\;#2%
}

\begin{document}

\title{Towards Human-Agent Communication via the Information Bottleneck Principle}


\author{\authorblockN{Mycal Tucker}\thanks{Appeared in \textit{Robotics Science and Systems Workshop on Social Intelligence in Humans and Robots}. New York City, USA 2022.}
\authorblockA{mycal@mit.edu}
\and
\authorblockN{Julie Shah}
\authorblockA{julie\_a\_shah@csail.mit.edu}
\and
\authorblockN{Roger Levy}
\authorblockA{rplevy@mit.edu}
\and
\authorblockN{Noga Zaslavsky}
\authorblockA{nogazs@mit.edu}}

\maketitle

\begin{abstract}
Emergent communication research often focuses on optimizing task-specific utility as a driver for communication. However, human languages appear to evolve under pressure to efficiently compress meanings into communication signals by optimizing the Information Bottleneck tradeoff between informativeness and complexity. In this work, we study how trading off these three factors --- utility, informativeness, and complexity --- shapes emergent communication, including compared to human communication. To this end, we propose Vector-Quantized Variational Information Bottleneck (VQ-VIB), a method for training neural agents to compress inputs into discrete signals embedded in a continuous space. We train agents via VQ-VIB and compare their performance to previously proposed neural architectures in grounded environments and in a Lewis reference game. Across all neural architectures and settings, taking into account communicative informativeness benefits communication convergence rates, and penalizing communicative complexity leads to human-like lexicon sizes while maintaining high utility. Additionally, we find that VQ-VIB outperforms other discrete communication methods. This work demonstrates how fundamental principles that are believed to characterize human language evolution may inform emergent communication in artificial agents.
\end{abstract}

\IEEEpeerreviewmaketitle

\section{Introduction}
Good communication is a critical component of successful teams of humans and artificial agents, but differing notions of ``good'' makes training agents to develop such communication challenging.
One view of communication focuses on \textit{utility} by framing languages as successful to the extent they enable high task performance.
Emergent communication literature, wherein agents learn to communicate while optimizing a task-specific reward or utility, emphasizes this view~\citep{Steels2005,foerster2016learning,lowe2017multi,mordatch2018emergence,lazaridou2016multi,lazaridou2020emergent}. However, this approach may lead to communication that is too complex for humans to understand~\citep{chaabouni2019antiefficient} or has slow convergence rates~\citep{eccles}, implying that additional constraints are needed in order to guide agents toward human-like communication. 

Here, we use constraints from cognitive science by building upon a recent line of work that argues that human languages evolve under pressure to efficiently compress meanings into communicative signals~\citep{Zaslavsky2018efficient,Zaslavsky2019containers,Zaslavsky2021person,Zaslavsky2022evolution,Mollica2021forms}. This notion of efficiency is formulated in terms of the Information Bottleneck (IB) principle~\citep{Tishby1999}, which is a general information-theoretic principle with broad scope in machine learning~\citep{Chechik2005,Shamir2010,Tishby2015DeepLA,alemi2016deep,Ziv2017}, and in this context, it can be interpreted as a tradeoff between the complexity and informativeness of communication~\citep{Zaslavsky2018efficient}. Intuitively, complexity corresponds to the number of bits that are needed for communication, and informativeness corresponds to how well a listener  can infer the speaker's intended meaning regardless of a specific task (e.g., humans understand what ``blue'' means regardless of any specific task, which may be ``avoid the blue box'' or ``find a blue cup'').
We hypothesize that taking into account complexity and informativeness, in addition to optimizing a task-specific utility function, may help guide artificial agents toward more natural, human-like communication systems.

\begin{figure*}[t!]
    \centering
    \begin{subfigure}[b]{0.28\textwidth}
        \centering
        \includegraphics[width=\textwidth]{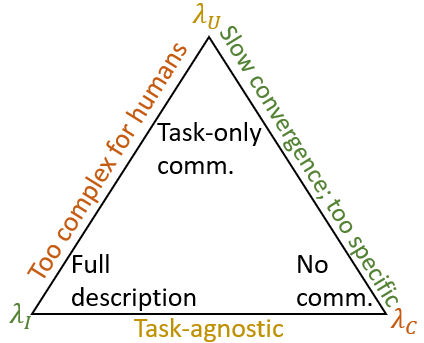}
        \caption{Theoretical Framework}
    \end{subfigure}
    \hfill
    \begin{subfigure}[b]{0.70\textwidth}
        \begin{overpic}[abs,unit=1mm, trim={0cm 0cm 7cm 0cm}, clip=true, width=0.40\textwidth]{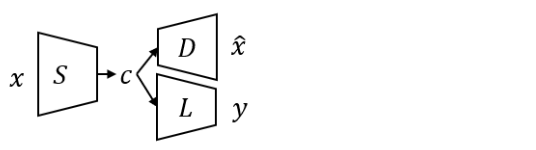}
        \put (55, 20) {Informativeness $:= -\mathbb{E}[d(X, \hat{X})]$}
        \put (55, 12) {Complexity $:= I(X, C)$}
        \put (55, 4) {Utility $:= U(X, Y)$}
        \end{overpic}
        \caption{Simplified Architecture. A speaker, $S$, maps an input, $x$, to a communication signal, $c$. A listener, $L$, takes an action, $y$, and a decoder, $D$, reconstructs $x$. Informativeness is negatively related to the expected distortion between $x$ and $\hat{x}$; complexity is defined as the mutual information between $X$ and $C$; utility can be any given task-specific reward function or loss.}
    \end{subfigure}
    \hfill
    \caption{Our theoretical framework and training losses allow us to weigh tradeoffs between utility ($\lambda_U$), informativeness ($\lambda_I$), and complexity ($\lambda_C$) to guide properties of emergent communication.}
    \label{fig:simplex}
\end{figure*}

We test this idea by training teams of artificial agents to optimize a tradeoff between maximizing task utility, maximizing communicative informativeness, and minimizing communicative complexity, as depicted in Figure~\ref{fig:simplex}. To this end, we extend prior work~\citep{vqvae,tucker2021emergent} and propose Vector-Quantized Variational Information Bottleneck (VQ-VIB), a method for training neural agents to compress inputs into signals, while maintaining a discrete~(symbolic) signal representation that is embedded in a continuous space. 
In experiments, we find that training to increase informativeness improves the convergence rate of emergent communication to high-performance communication, while annealing the complexity loss generates a spectrum of learned communication, which we show aligns with a variety of human languages.
Additionally, we find that VQ-VIB appears better able to achieve high utility for the same complexity and informativeness than other discrete communication methods.
This suggests a promising avenue for supporting human-robot communication by instilling inductive biases inspired by cognitive science.

\section{Related Work}
\label{related}
Here, we describe related work in emergent communication, including in a specific color domain which we use; we include approaches for measuring informativeness and complexity in Section~\ref{sec:technical}.
Generally, in emergent communication settings, researchers train agents to accomplish a cooperative task in which communication should enable better task performance~\citep{lowe2017multi,lazaridou2016multi,bullard2020exploring,bullard2021quasi,lazaridou2018emergence}.
Some works simplify communication by discretizing communication, by adding noise to the environment, or passing communication through a centralized message processor; these methods all correspond to limiting complexity~\citep{tucker2021emergent,catalytic,imac}.
At the same time, other works find that auxiliary losses designed to create ``meaningful'' communication improve convergence rates~\citep{eccles,jaques2019social,marlae}, often inspired by notions adjacent to informativeness.
Color reference games have been widely used in emergent communication literature \citep{Steels2005, Dowman2007,Kageback2020rl,Chaabouni2021communicating}, mirroring the centrality of color naming for studying humans' semantic categorization and language evolution~\citep{Berlin1969}.
Most relevant to our work is \citet{Zaslavsky2018efficient}'s finding that color naming systems across languages (from a large-scale, cross-linguistic dataset, the Word Color Survey~\citep[WCS;][]{Kay2009}) attain complexity--informativeness tradeoffs that are near-optimal in the Information Bottleneck~\citep[IB;][]{Tishby1999} sense. \citet{Chaabouni2021communicating} further showed that AI agents can learn IB-efficient color naming systems by playing color reference games (and see~\citep{Kageback2020rl} for a related study using different notions of efficiency).
While emergent communication literature has begun to rediscover the importance of complexity and informativeness in communication, these terms have rarely been explicitly recognized, whereas we include terms for utility, informativeness, and complexity in training agents.

\section{Technical Approach}
\label{sec:technical}
Here, we show how to adapt standard neural architectures to enable informativeness- and complexity-based objectives.
We then introduce a new neural architecture, Vector-Quantized Variational Information Bottleneck (VQ-VIB), for learning complexity-limited discrete representations in a continuous space.
Lastly, we combine utility, informativeness, and complexity terms into an overall objective.

\subsection{Information Bottleneck for Semantic Communication}

We adopt and extend~\citet{Zaslavsky2018efficient}'s widely-supported IB framework for semantic communication~\citep{Zaslavsky2019containers,Zaslavsky2021person,Zaslavsky2022evolution,Mollica2021forms}. In this framework, speakers and listeners jointly optimize the IB tradeoff between the complexity and informativeness of their communication system. A speaker is characterized as a probabilistic encoder $q(c|x)$ that, given an input, $x\sim p(x)$, generates a signal, $c$. A listener is characterized as a probabilistic decoder $q(\hat{x}|c)$, that given $c$ aims to reconstruct the speaker's input.
Complexity is measured by the mutual information between the speaker's inputs and signals, $I(X, C)$. Informativeness is measured by the degree to which the speaker's mental representation of $x$ matches the listener's mental representation of $\hat{x}$. In general, these mental representations are defined by probability distributions, $m$ and $\hat{m}$ respectively, over some feature space. In that case, minimizing the expected Kullback-Leibler (KL) divergence $\mathbb{E}[D[m\|\hat{m}]]$ amounts to maximizing the informativeness of the communication. This yields the following optimization principle: $\min I(X,C)-\beta \mathbb{E}[D[m\|\hat{m}]]$, where $\beta$ is a tradeoff parameter. It has been shown that this principle is equivalent to the IB principle~\citep{Zaslavsky2018efficient}.

Solving this optimization problem directly is challenging in large scale domains and with agents that are parameterized by neural networks. To address this, we take a Variational Information bottleneck~\citep[VIB;][]{alemi2016deep} approach, in which an upper bound is optimized, similar to ELBO in Variational Autoencoders~\citep[VAEs,][]{vae} and $\beta$-VAEs~\citep{betavae}.
We also approximate informativeness by an autoencoding loss, $d(x,\hat{x})$, defined by the mean squared error (MSE) loss of the speaker's input and the listener's reconstruction.

\subsection{Vector Quantized Variational Information Bottleneck}
\label{sec:vqvib}
Our proposed neural architecture, the Vector Quantized Variational Information Bottleneck (VQ-VIB), learns information-constrained discrete communication in a continuous space.
Our technique combines prior work in discrete representation learning~\citep[VQ-VAE,][]{vqvae} with $\beta$-VAE~\citep{betavae}.

\subsubsection{Vector-Quantized Variational Autoencoder}
First, consider a simplified Vector-Quantized Variational Autoencoder (VQ-VAE) architecture~\citep{vqvae}.
As in standard VAEs, an encoder is a conditional distribution over latent representations, given an input: $q(z|x)$.
A deterministic network maps from an input, $x$, to a latent, $z$: we dub such encodings, $z(x)$.
Next, for a VQ-VAE parametrized with $K$ embedding vectors, $\boldsymbol{\zeta}$, in the latent space, $z(x)$ is discretized to a particular embedding vector, $\zeta_i$, by mapping to the nearest embedding vector.

\subsubsection{$\beta$ Variational Autoencoder ($\beta$-VAE)}
We also use notions from $\beta$-VAE, a variational approach for generating information-constrained continuous representations with neural nets.
Following \citet{betavae}, an encoder network maps from an input to $\mu$ and $\sigma$, which are used as parameters to a Gaussian distribution from which a latent representation, $z$, is sampled.
$\beta$-VAE networks are trained via a weighted combination of a reconstruction loss and ($\beta$ times) the KL divergence of $\mu(x)$ and $\sigma(x)$ from a unit Gaussian; this KL loss corresponds to an upper bound of $I(X, Z)$.
\begin{figure}[t!]
    \centering
    \begin{subfigure}[b]{0.45\textwidth}
        \centering
        \includegraphics[width=\textwidth]{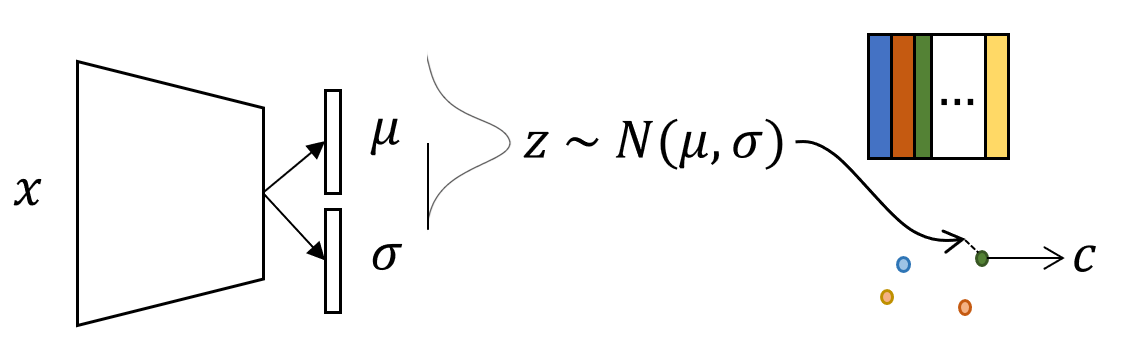}
    \end{subfigure}
    \caption{VQ-VIB architecture blends Vector Quantization and Variational Information Bottleneck. The speaker encodes an input, $x$, as parameters to a Gaussian, from which a communication vector is sampled and then discretized by mapping to the nearest encoding, to produce communication, $c$.}
    \label{fig:vqvib_arch}
\end{figure}

\subsubsection{VQ-VIB Architecture}
Here, we integrate VQ-VAE and $\beta$-VAE for a method of generating discrete, information-bounded representations.
Our architecture, Vector-Quantized Variational Information Bottleneck (VQ-VIB), corresponds to first using an encoder to sample a representation $z$ in a continuous space, followed by a quantization layer that discretizes the sample by mapping to the nearest encoding $\zeta$.
This is depicted in Figure~\ref{fig:vqvib_arch} and is formally defined in Equation~\ref{eq:vqvib_sb}.
The likelihood of a communication vector, $c$, given an input, $x$, corresponds to the likelihood that an encoding $z$, sampled with parameters $\mu(x)$ and $\sigma(x)$, is closest to an embedding vector equal to $c$ (Eq.~\ref{eq:vqvib_sb}).
This intuitively corresponds to notions of uncertainty over which ``word'' to use.
One may compute an upper bound on $I(X, C)$ based on this uncertainty by using the standard KL divergence used for Gaussians in a $\beta$-VAE, using $\mu(x)$ and $\sigma(x)$.

\begin{equation}
    \label{eq:vqvib_sb}
    \begin{split}
        &z \sim \mathcal{N}(\mu(x), \sigma(x))\\
        &q(c|x) = \mathbb{P}(c = \text{argmin}_{\zeta} [(z - \zeta)^2]\\
    \end{split}
\end{equation}

\begin{figure}[tb]
    \centering
    \begin{subfigure}[b]{0.23\textwidth}
        \centering
        \includegraphics[width=\textwidth,height=3.0cm]{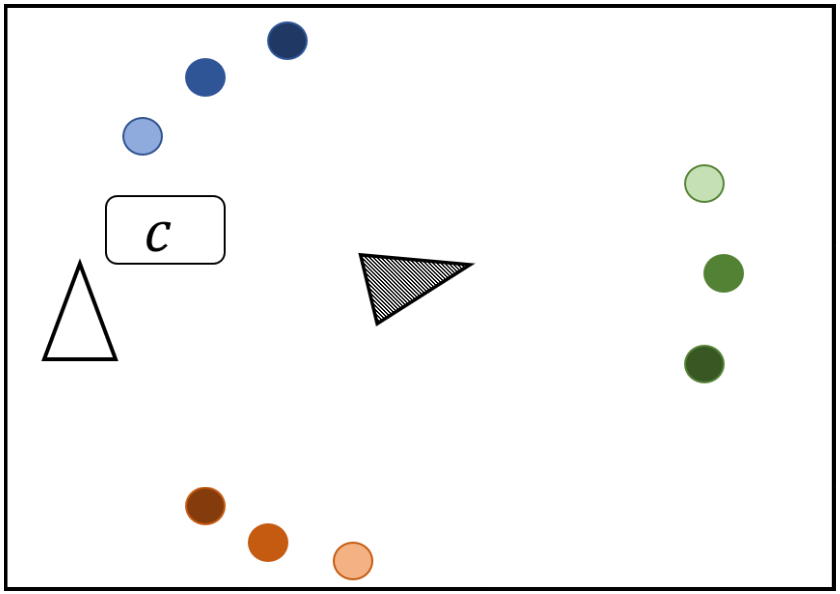}
        \caption{9 Points}
    \end{subfigure}
    ~
    \begin{subfigure}[b]{0.23\textwidth}
        \centering
        \includegraphics[width=\textwidth,height=3.0cm]{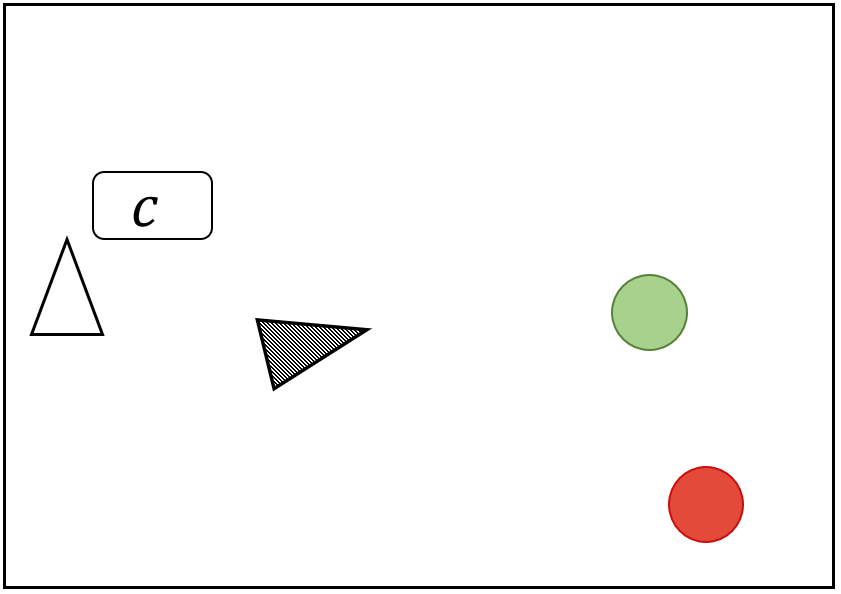}
        \caption{Uniform}
    \end{subfigure}
    \caption{In particle-world environments, a speaker (clear) communicated to a listener (gray) to reach a desired landmark.}
    \label{fig:envs}
\end{figure}

\begin{figure}[t]
    \begin{subfigure}[b]{0.48\textwidth}
        \includegraphics[trim={2cm 11.0cm 0cm 1cm}, clip=true, width=\textwidth,height=1.7cm]{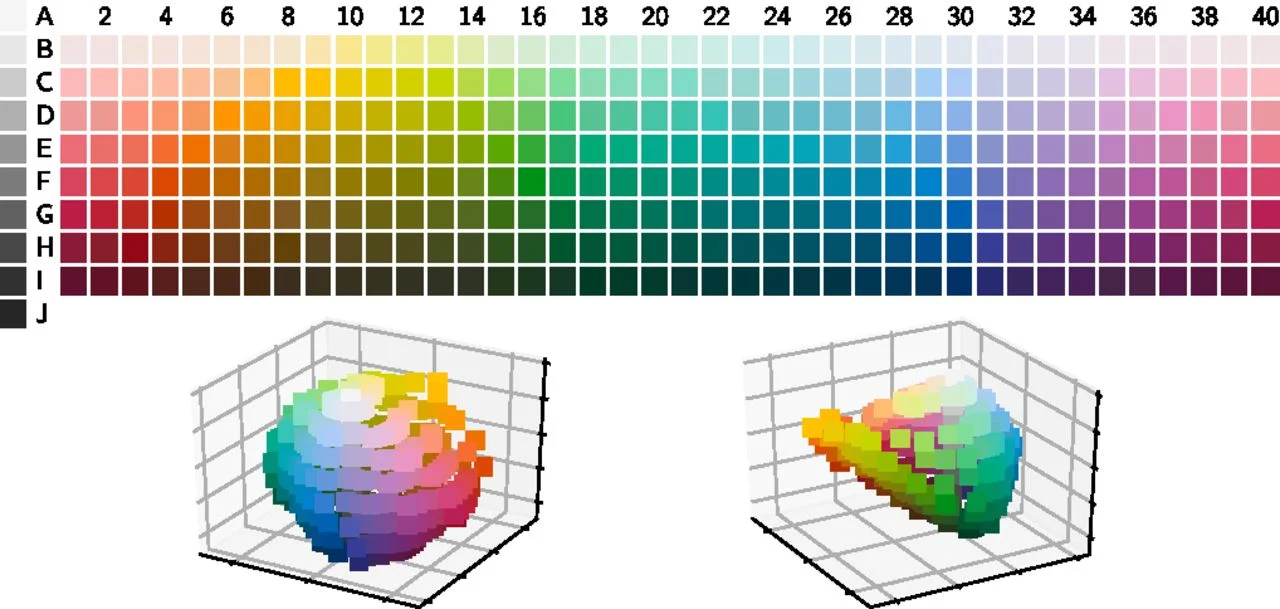}
    \end{subfigure}
    \caption{In the color reference game, agents observed colors from the WCS Dataset~\citep{Kay2009}.
    }
    \label{fig:color_env}
\end{figure}

\begin{figure}[ht]
    \centering
    \begin{subfigure}[c]{0.45\textwidth}
        \includegraphics[width=\textwidth]{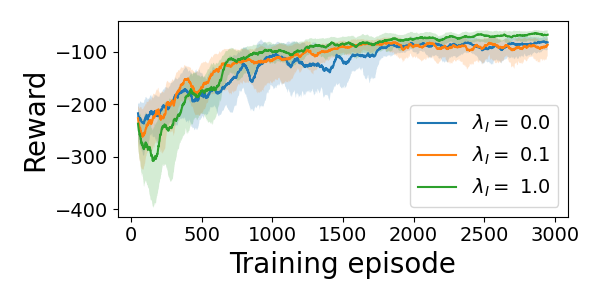}
    \end{subfigure}
    \begin{subfigure}[c]{0.45\textwidth}
        \includegraphics[width=\textwidth]{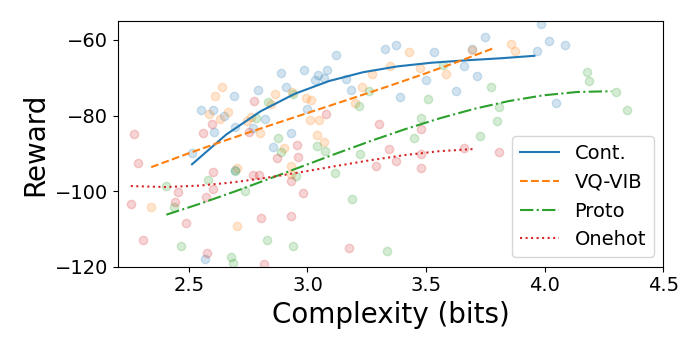}
    \end{subfigure}
    \caption{\textbf{Top:} In the 9 points environment, greater $\lambda_I$ led to faster convergence for VQ-VIB; \textbf{Bottom}: Annealing $\lambda_C$ decreased reward and complexity for all methods.}
    \label{fig:9points_results}
\end{figure}

\begin{figure*}[t!]
    \centering
    \begin{subfigure}[c]{\textwidth}
        \centering
        \includegraphics[width=\textwidth]{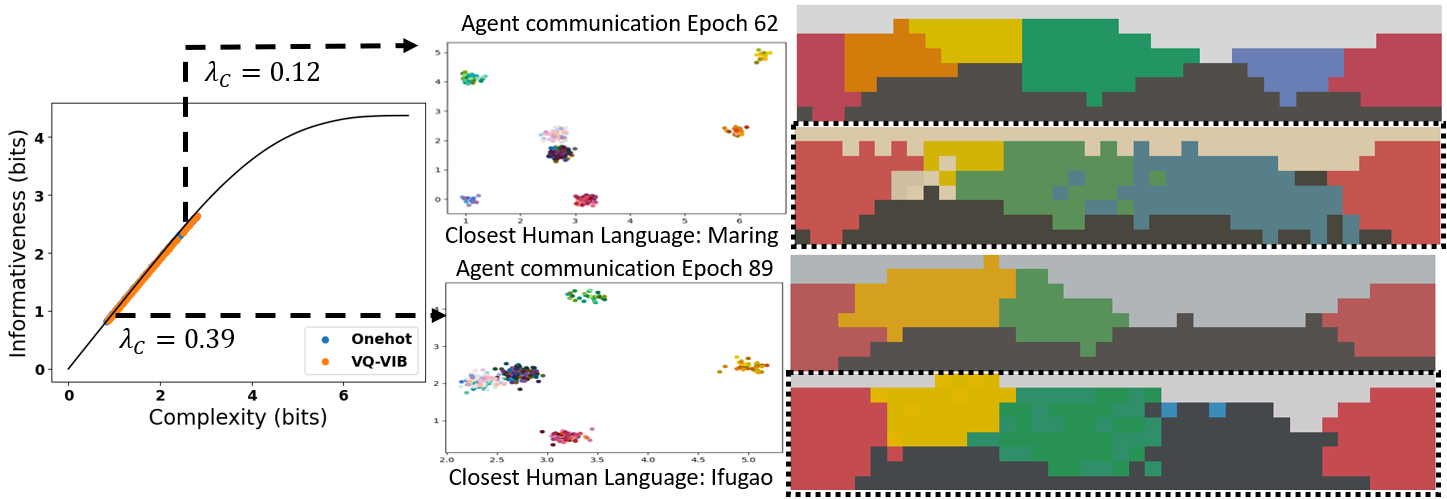}
    \end{subfigure}
    \caption{Color reference game results. Left: By annealing complexity, we recreated a range of near-optimal color-naming schemes (black curve is the theoretical IB bound from~\citep{Zaslavsky2018efficient}). Middle: 2D PCA of VQ-VIB Before communication shows relationships between different clusters in the communication space.
    Right: mode maps of the agent (no box) and human (dashed box) color naming systems, where each color chip from \Cref{fig:color_env} is colored by the centroid of its modal category.}
    \label{fig:isolated_results}
\end{figure*}

\subsection{Overall Optimization}
Here, we combine losses into an idealized objective function to maximize, shown in Equation~\ref{eq:ideal}.
This objective function is the weighted combination of three terms, wherein we seek to maximize a tradeoff between high utility, high informativeness (low expected distortion $d$), and low complexity.
Intuitively, these three scalar weights form a simplex that describe a range of communication from silence (minimizing complexity) to task-optimized communication that does not generalize (maximizing utility) to complete descriptions (maximizing informativeness).

\begin{equation}
    \label{eq:ideal}
    \text{maximize} \quad \lambda_U U(X,Y) - \lambda_I \mathbb{E}[d(X, \hat{X})] - \lambda_C I(X, C)
\end{equation}

In practice, we train neural agents via gradient descent to maximize approximations of these terms.
Utility is represented via the policy gradient (for RL settings) or a supervised loss.
Informativeness is computed via the negative MSE between the speaker's input and a decoder's output, given the communication.
Lastly, complexity is bounded by the KL divergence between the speaker's distribution over communication and a fixed marginal (either a uniform distribution for onehot communication, or a unit Gaussian for others).

\section{Results}

\subsection{Particle Worlds}
We trained agents in two types of 2D particle worlds, as depicted in Figure~\ref{fig:envs}.
In each environment, the speaker observed a ``target'' landmark (in 9 points, there were 9 landmarks at fixed locations, whereas in Uniform, the landmarks were spawned a random locations) and communicated to the listener.
The team reward was the negative distance from the listener to the target, so agents learned to communicate about the target location.
We trained agents with Multi-Agent Deep Deterministic Policy Gradient~\citep[MADDPG;][]{lowe2017multi} and compared continuous (real-valued), onehot, VQ-VIB, and prototype-based~\citep[Proto;][]{tucker2021emergent} communication.
All except continuous communication are fundamentally discrete.
For all environments and architectures, training agents with higher $\lambda_I$ improved training convergence, as depicted for VQ-VIB in the 9-points environment in Fig.~\ref{fig:9points_results} (top).
Furthermore, as we increased $\lambda_C$, all architectures learned less complex, but also lower-utility, communication (Fig.~\ref{fig:9points_results} bottom).
However, VQ-VIB achieved higher utility, for the same complexity, than other discrete communication methods.
These trends also appeared in the Uniform environment: 1) higher $\lambda_I$ improved convergence, 2) increasing $\lambda_C$ created a spectrum of communication complexity and 3) VQ-VIB outperformed other discrete methods.

\subsection{Color Reference Games}
We also trained agents in a color reference game in which a speaker agent observed a target color (drawn from WCS data, as shown in Fig.~\ref{fig:color_env}), and a listener agent had to distinguish between the target and a distractor color.
As in the previous experiments, we found that increasing $\lambda_I$, for a fixed $\lambda_C$, improved convergence rates, and VQ-VIB agents typically learned more complex communication than onehot agents.

We tested agents for $\lambda_I=1.0$ while linearly annealing $\lambda_C$ to penalize complexity.
As $\lambda_C$ increased, the communication complexity decreased (Figure~\ref{fig:isolated_results} left).
Over 5 random seeds, the points in the plot, generated at each training epoch, closely approximated the IB upper bound for maximum informativeness for a given complexity.
The learned color naming schemes at different complexity levels exhibited important patterns (Figure~\ref{fig:isolated_results} right).
As complexity decreased, the number of communication clusters emitted by VQ-VIB agents decreased (visualized via 2-dimensional principle component analysis).
Clustering in the communication space was reflected in the color space (e.g., ``yellow'' and ``orange'' merging).
More generally, a least-squares linear correlation between the distance between pairs of communication vectors and the distance between associated colors (in perceptually-based CIELAB space) showed $r^2 = 0.97 \pm (0.01)$ for VQ-VIB, whereas for onehot communication $r^2 = 0.76 \pm (0.03)$.
That is, for VQ-VIB, if two communication vectors were far apart in communication space, they described colors that were far apart.
This shows a fundamental connection between the form and meaning of communication used by VQ-VIB agents, which onehot-based communication is architecturally unable to support.

Lastly, we compared the agent color naming schemes to human language data.
For each of the 110 languages in the WCS Dataset \citep{Kay2009}, we found the best-matching agent as determined by gNID, a measure of dissimilarity of naming distributions introduced by \citet{Zaslavsky2018efficient}.
Example nearest languages are shown in Figure~\ref{fig:isolated_results} in dashed boxes.
Annealing complexity of artificial communication was necessary to generate similar naming conventions for each human language.
Furthermore, while onehot and VQ-VIB Before achieved similar informativeness for the same complexity, VQ-VIB Before reached higher utility, indicating an architectural advantage.

\section{Conclusion} 
\label{sec:conclusion}
In this work, we explore the interplay of utility, informativeness, and complexity in emergent communication.
To the best of our knowledge, we are the first to directly optimize agents to communicate according to a combination of all three terms.
Rewarding informativeness led to faster communication convergence, and penalizing complexity created more human-like communication.
While we focused on training agents entirely in ``self-play,'' natural extensions of this work would include supporting human-robot communication, wherein robots could learn partner-specific communication by appropriately constraining complexity and a small amount of supervised data.

\section*{Acknowledgments}
M.T. is supported by an Amazon Science Hub Fellowship. N.Z. is supported by a K. Lisa Yang Integrative Computational Neuroscience (ICoN) Postdoctoral Fellowship.


\bibliography{references}

\end{document}